\documentclass{article}

\usepackage{arxiv}

\usepackage[utf8]{inputenc} % allow utf-8 input
\usepackage[T1]{fontenc}    % use 8-bit T1 fonts
\usepackage{hyperref}       % hyperlinks
\usepackage{url}            % simple URL typesetting
\usepackage{booktabs}       % professional-quality tables
\usepackage{amsfonts}       % blackboard math symbols
\usepackage{nicefrac}       % compact symbols for 1/2, etc.
\usepackage{microtype}      % microtypography
\usepackage{lipsum}
\usepackage{fancyhdr}       % header
\usepackage{graphicx}       % graphics
\graphicspath{{media/}}     % organize your images and other figures under media/ folder

\usepackage[numbers,super,sort&compress]{natbib}

\usepackage{doi}
\usepackage{dashrule}
\usepackage{markdown}
\usepackage{graphicx}
\usepackage{enumitem}
\usepackage{xcolor}
\usepackage{enumitem,amssymb}
\newlist{todolist}{itemize}{2}
\setlist[todolist]{label=$\square$}
\usepackage{pifont}
\usepackage{xcolor}   % For colored symbols
\usepackage{soul}
\usepackage{capt-of}

\usepackage{float}

\usepackage{lineno}
\usepackage[skins]{tcolorbox}

\usepackage{tikz}
\usetikzlibrary{positioning,arrows.meta, calc}

\usepackage{setspace}
\renewcommand\hl[1]{#1} 

\graphicspath{ {images/} }

\newcommand{\correctbox}{\rlap{$\square$}{\kern0.2em\textcolor{green}{\ding{51}}}}  % Correct: Green Checkmark
\newcommand{\incorrectbox}{\rlap{$\square$}{\kern0.2em\textcolor{red}{\ding{55}}}}  % Incorrect: Red X
\definecolor{MyBlue}{rgb}{0.6,0.6,0.8}
\title{Advances in LLM Reasoning Enable Flexibility in Clinical Problem-Solving}

\author{ 
    \textbf{Kie Shidara} \\
	Weill Institute of Neurology and Neurosciences\\
	University of California, San Francisco\\
	\and
    \textbf{Preethi Prem} \\
    Carle Illinois College of Medicine\\
	University of Illinois Urbana-Champaign\\
	\and
    \textbf{Jonathan Kim} \\
	Department of Neurology and Neurological Sciences\\
	Stanford University\\
    \and
    \textbf{Anna Podlasek} \\
    Image Guided Therapy Research Facility \\
    University of Dundee \\
    \and
    \textbf{Feng Liu} \\
    Department of Systems Engineering \\
    Stevens Institute of Technology \\
    \and
	\textbf{Ahmed Alaa} \\
	Department of EECS\\
	University of California Berkeley\\
	\and
        \textbf{Danilo Bernardo} \\
	Weill Institute of Neurology and Neurosciences\\
	University of California, San Francisco\\
	\texttt{dbernardoj@gmail.com} \\
}

\hypersetup{
pdftitle={LLM Reasoning Advances Enable Flexibility in Clinical Problem-Solving},
pdfsubject={q-bio.NC, q-bio.QM},
pdfauthor={Kie Shidara},
}
\begin{document}
\maketitle

\begin{abstract}
Large Language Models (LLMs) have achieved high accuracy on medical question–answer (QA) benchmarks, yet their capacity for flexible clinical reasoning has been debated. Here, we asked whether advances in reasoning LLMs improve their cognitive flexibility in clinical reasoning. We assessed reasoning models from the OpenAI, Grok, Gemini, Claude, and DeepSeek families on the medicine abstraction and reasoning corpus (mARC), an adversarial medical QA benchmark which utilizes the Einstellung effect to induce inflexible overreliance on learned heuristic patterns in contexts where they become suboptimal. We found that strong reasoning models avoided Einstellung-based traps more often than weaker reasoning models, achieving human-level performance on mARC. On questions most commonly missed by physicians, the top 5 performing models answered 55\% to 70\% correctly with high confidence, indicating that these models may be less susceptible than humans to Einstellung effects. Our results indicate that strong reasoning models demonstrate improved flexibility in medical reasoning, achieving performance on par with humans on mARC.
\end{abstract}

% keywords can be removed
% \keywords{LLM \and AI \and medicine \and reasoning}

\section{Introduction}
The versatility and strong performance of Large Language Models (LLMs) across multiple domains\cite{bubeck2023sparks} have motivated their assessment in clinical contexts\cite{moor2023foundation}. High scores have been reported on the United States Medical Licensing Exam (USMLE)\cite{gilson2023does}, USMLE‑styled question banks\cite{jiang2023health, peng2023study, zhang2024generalist}, subspecialty board exams\cite{longwell2024performance, schubert2023performance}, and validated clinical reasoning benchmarks\cite{cabral2024clinical}. These results have been linked to emergent reasoning abilities\cite{lievin2024can, kwon2024large}, yet performance in realistic, novel scenarios remains the crucial test\cite{hager2024evaluation, williams2024evaluating}. 

While LLMs display impressive factual recall and pattern recognition, evidence shows that their reasoning abilities, particularly in medical contexts, remain poor\cite{kim2025limitations}. Studies have shown that when used for real-life clinical tasks requiring reasoning\cite{valmeekam2024planbench} or abstraction\cite{mitchell2023comparing}, LLMs often produce inconsistent, overconfident, and logically flawed responses\cite{griot2025large}. Despite succeeding on medical board-style questions, these models struggle to generalize beyond familiar exam patterns, failing in settings that require flexible rationalization, such as determining admission status, radiological investigation(s) request status, and antibiotic prescription status\cite{williams2024evaluating}. This highlights that these limitations arise from an overreliance on surface-level correlations within the training data rather than genuine reasoning \cite{griot2025large}. LLMs inability to plan\cite{valmeekam2024planbench}, reason under uncertainty\cite{kim2025limitations}, or reflect metacognitively\cite{griot2025large} leads to reasoning failures that are masked by superficially fluent and overconfident language generation. This overconfidence in incorrect answers poses unique risks in medicine, where diagnostic reasoning depends on contextual synthesis rather than rote memorization\cite{durning2016functional}. The literature on LLMs converges on a key insight that current LLMs may imitate medical reasoning without underlying cognitive flexibility, thus calling into question their capabilities and safety in real-world medical scenarios\cite{griot2025large}, and necessitating targeted evaluation of LLM reasoning capabilities.

More recently, however, LLMs with \emph{reasoning} of operation have emerged which utilize detailed multi-step reasoning chains, reminiscent of human-like \emph{thinking} about a problem before answering\cite{marjanovic2025deepseek}. Reasoning models have demonstrated gradual performance improvements in reasoning-based benchmarks such as Francois Chollet's Abstraction and Reasoning Corpus (ARC) challenge \cite{chollet2024arc}, suggesting that they have improved generalized reasoning abilities which may potentially translate to improved performance at medical reasoning. We previously introduced a novel medical QA evaluation, medical abstraction and reasoning corpus (mARC), as a probe to assess flexibility in medical reasoning by stress-testing the \emph{Einstellung} effect, a cognitive bias in which reliance on familiar patterns inhibits the discovery of better solutions\cite{luchins1942mechanization, bilalic2010mechanism}. mARC problems were designed to elicit the \textit{Einstellung} effect, rigid, heuristic completions triggered by familiar cues, and led to inflexible reasoning by LLMs\cite{kim2025limitations}. Here, we assess whether newer reasoning models may overcome \textit{Einstellung} effect in mARC. 

\begin{figure}[ht]
\begin{flushleft}
\begin{tikzpicture}[
    node distance=1cm and 2cm,
    box/.style={
        rectangle,
        draw=black!60,
        rounded corners,
        align=left,
        minimum width=2.5cm,
        minimum height=1cm,
        font=\sffamily\small,
        fill=white,
        thick
    },
    arrow/.style={->, >=Stealth, thick},
    dashed_arrow/.style={->, >=Stealth, thick, dashed},
    label_text/.style={font=\footnotesize\itshape, text=black!70}
]

% --- Top Level: The Input ---
\node[box, fill=gray!10, minimum width=8cm, align=left, anchor=west] (vignette) {
    \textbf{Clinical Vignette (Input)} \\[0.25em]
    \emph{Patient $p$ presents with reduced alertness ($S$)} \\
    Cue $C$: Anticoagulation \hspace{1em} Blocker $B$: Anencephaly
};

% --- Explicit Background Knowledge K (moved to top level) ---
\node[box,
      fill=blue!2,
      right=1cm of vignette,
      minimum width=4cm,
      align=left] (knowledge) {
    \textbf{Background medical knowledge $K$}\\[0.25em]
    1.\ $\text{BrainBleed}(x)\rightarrow\text{BrainPresent}(x)$\\
    2.\ $\text{Anencephaly}(x)\rightarrow\neg\text{BrainPresent}(x)$
};

% --- Left Branch: The Heuristic Trap ---
\node[box, fill=red!5, below left=of vignette, xshift=7cm] (cue_process) {
    \textbf{Default Heuristic} \\[0.25em]
    $C \Rightarrow_{\text{def}} H$ \\
    (Anticoag $\to$ Bleed)
};

\node[box, fill=red!10, below=of cue_process, draw=red!50] (option_b) {
    \textbf{Adversarial Option} \\[0.25em]
    ``Obtain CT Head'' \\
    \emph{Failure to suppress default heuristic}
};

% --- Right Branch: The Logical Constraint from K ---
\node[box, fill=blue!5, below right=of vignette, xshift=-2cm] (blocker_process) {
    \textbf{Deductive Constraint from $K$ and $B$} \\[0.25em]
    $K \cup \{B\} \vdash \neg H$ \\
    (Anencephaly $\to$ No Brain $\to$ No Bleed)
};

\node[box, fill=blue!10, below=of blocker_process, draw=blue!50] (option_d) {
    \textbf{Correct Option} \\[0.25em]
    ``Obtain More Clinical Info'' \\
    \emph{Successful override of default heuristic}
};

% Arrow from K to the deductive constraint (updated path)
\draw[arrow]
  (knowledge.south) -- ($(knowledge.south |- blocker_process.north)$)
  node[midway, right, label_text] {used with $B$};

% --- Arrows and Interactions from the Vignette ---
\draw[arrow] ([xshift=3cm]vignette.south) -- ++(0,-0.5)
    -| (cue_process.north)
    node[pos=0.25, above, label_text] {activates};

\draw[arrow] ([xshift=3cm]vignette.south) -- ++(0,-0.5)
    -| (blocker_process.north);

% The Default Path
\draw[arrow] (cue_process) -- (option_b);

% The Logical Path
\draw[arrow] (blocker_process) -- (option_d);

% The Conflict / Override
\draw[dashed_arrow, red, ultra thick] (blocker_process) -- (cue_process)
    node[midway, above, font=\bfseries\footnotesize, fill=white, inner sep=1pt] {BLOCKS};

% --- Background Annotations ---
\node[below=0.15cm of option_b, text width=4cm, align=center]
    {Driven by susceptibility to \emph{Einstellung} effect};

\node[below=0.15cm of option_d, text width=4cm, align=center]
    {Driven by reasoning over background knowledge $K$};

\end{tikzpicture}
\caption{\textbf{Demonstration of \emph{Einstellung} effect evoked by failure to override default heuristics.}
The vignette presents a cue ($C$) that triggers a default heuristic for brain bleed ($H$), and a blocker ($B$) that, together with background medical knowledge $K$ (e.g., that intracranial hemorrhage requires a brain and anencephaly entails absence of brain), entails $\neg H$.
An \emph{Einstellung} effect type reasoning failure occurs when the model prioritizes the default heuristic $C \Rightarrow_{\text{def}} H$ over the deductive constraint $K \cup \{B\} \vdash \neg H$.}
\label{fig:reasoning_conflict}
\end{flushleft}

\end{figure}
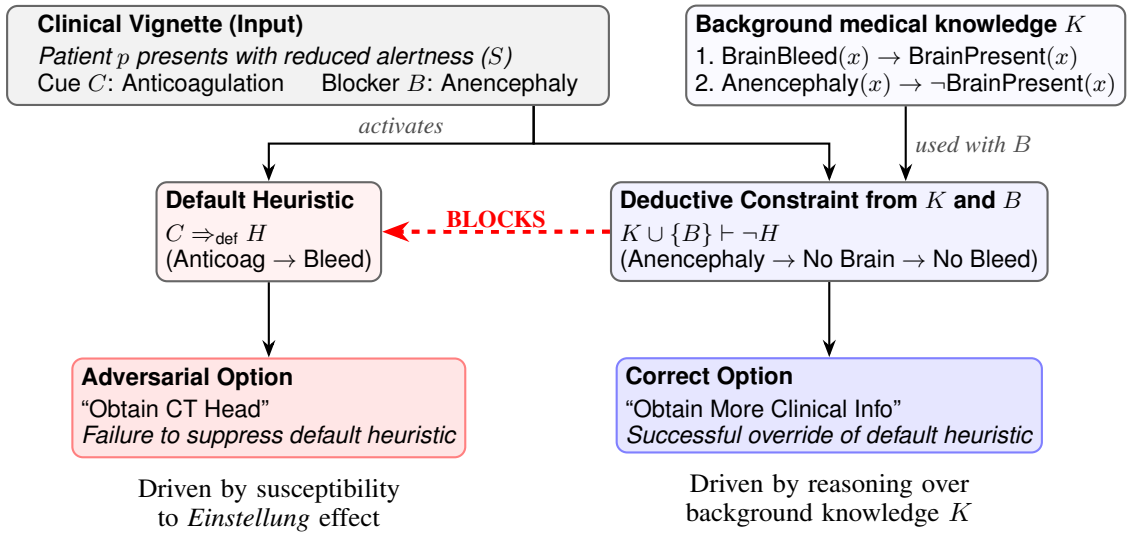

\section{Methods}

\emph{Medicine Abstraction and Reasoning Corpus (mARC) Dataset.} mARC questions follow the USMLE‑style multiple‑choice format and comprises 100 author‑written multiple choice questions crafted to resist rote pattern matching or interpolation from existing medical QA corpora, which were previously validated by board-trained specialists\cite{kim2025limitations}. We instantiate the \textit{Einstellung} effect, rigidity of thought elicited by familiar problem cues\cite{bilalic2010mechanism, luchins1942mechanization}, as adversarial design elements and measure whether test-takers can override these pressures. Fifty‑three percent of questions include ``seek more data’’ to test adaptive deferral. Subspecialties span neurology, neurosurgery, infectious disease, obstetrics‑gynecology, ophthalmology, HEENT, hematology‑oncology, gastroenterology, pulmonology, critical care, cardiology, and emergency medicine. To isolate flexibility from knowledge gaps, items require at most broadly taught, early‑clinical knowledge that were previously validated as appropriate for medical‑school graduates\cite{kim2025limitations}.

\emph{Operationalizing the Einstellung effect.} Each item triggers a stereotyped heuristic while making that heuristic counterproductive; correct solutions require cognitive flexibility. We demonstrate  formalization of the \textit{Einstellung} effect demonstrating cue conflict with decisive blockers in Figure 1, with detailed description below.

\begin{enumerate}[leftmargin=*, nosep]

     \item {Knowledge base and default rule.}
    Background medical knowledge $K$ was represented as:
    \[
    \text{BrainBleed}(x) \rightarrow \text{BrainPresent}(x), \qquad
    \text{Anencephaly}(x) \rightarrow \neg \text{BrainPresent}(x),
    \]
    yielding:
    \[
    K \vdash \text{Anencephaly}(x) \rightarrow \neg \text{BrainBleed}(x),
    \]
    i.e., an anencephalic patient cannot have an intracranial hemorrhage.

    Clinicians were assumed to hold a cue-based default rule:
    \[
    \text{Anticoag}(x) \Rightarrow_{\text{def}} \text{BrainBleed}(x),
    \]
    capturing the stereotyped concern for brain bleed in anticoagulated patients.\\
    
     \item {Vignette construction (cue conflict with decisive blocker).}
    We constructed a clinical vignette for a patient $p$ that instantiated:
    \[
    B := \text{Anencephaly}(p), \qquad
    C := \text{Anticoag}(p),
    \]
    together with a non-specific symptom $S$ (reduced alertness) compatible with multiple etiologies, including brain bleed. Under $K$, $B$ is a \emph{decisive blocker} for
    \[
    H := \text{BrainBleed}(p),
    \]
    since:
    \[
    K \cup \{B\} \vdash \neg H
    \quad\text{and}\quad
    K \cup \{B, H\} \text{ is inconsistent.}
    \]

     \item {Response options and coding.}
    Two options are provided amongst answer selections:
    \begin{itemize}[leftmargin=4cm, rightmargin=3cm]
        \item[{\emph{Einstellung} Trap:}] \emph{Obtain CT head} — unrevised application of the default heuristic $C \Rightarrow_{\text{def}} H$ despite the blocker $B$.
        \item[{Correct Reasoning:}] \emph{Obtain more clinical information} — overriding the default heuristic in light of $K \cup \{B\} \vdash \neg H$ and seeking alternative explanations for $S$.
    \end{itemize}

\end{enumerate}

\emph{Data Collection and Analysis.} We compared LLM performance to physician performance on mARC. Physicians (five total; specialties included pediatrics, internal medicine, and neurology) were recruited from UCSF Medical Center and kolabtree.com. \hl{Ethical approval was obtained from the UCSF IRB (IRB\#24‑42911); informed consent was obtained from all participants; all experiments complied with relevant guidelines.} Physicians took the exam online with a 2‑hour limit; the reported human score is the mean of five physician accuracies.

Models assessed included OpenAI GPT‑4o\cite{hurst2024gpt}, OpenAI o1\cite{jaech2024openai}, OpenAI 5.1, Claude‑Sonnet/Opus\cite{anthropic2024claude3}, Google Gemini \cite{team2023gemini}, Mistral\cite{jiang2023mistral}, DeepSeek R1 and V3\cite{guo2025deepseek}, Llama \cite{touvron2023llama}, and Grok\cite{xai_grok}; closed‑source models were evaluated via provider APIs, open‑source via HuggingFace/Lambda Labs. Chain‑of‑thought (CoT) prompting followed MMLU‑Pro methodology\cite{hendrycks2020measuring, wang2024mmlu}; temperature was 0 when possible; otherwise defaults followed Wang et al.\cite{wang2024mmlu}. 

Uncertainty estimation used agreement-based and entropy-based sample‑consistency (SC)\cite{xiong2023can, wang2022self, lyu2024calibrating}, which outperforms token probability and elicitation in the medical domain\cite{savage2025large}. To induce stochasticity across runs, we varied patient age by up to 10 days (no neonatal/infant cases). We constructed reliability plots and Brier scores following Lyu et al.\cite{lyu2024calibrating}. We report average accuracy, sample- and entropy-based consistency (SC and EC) across 15 runs per model (plateauing of these measures has been reported beyond this sample size\cite{manakul2023selfcheckgpt}). Confidence intervals were computed by bootstrap (2{,}000 resamples) with Benjamini–Hochberg correction. The mARC problem dataset and analysis code are available at \url{https://github.com/bernardolab/mARC-Reasoning}. 

\section{Results}

\begin{figure}[!htbp]
	\centering
        \includegraphics[width=1\textwidth]{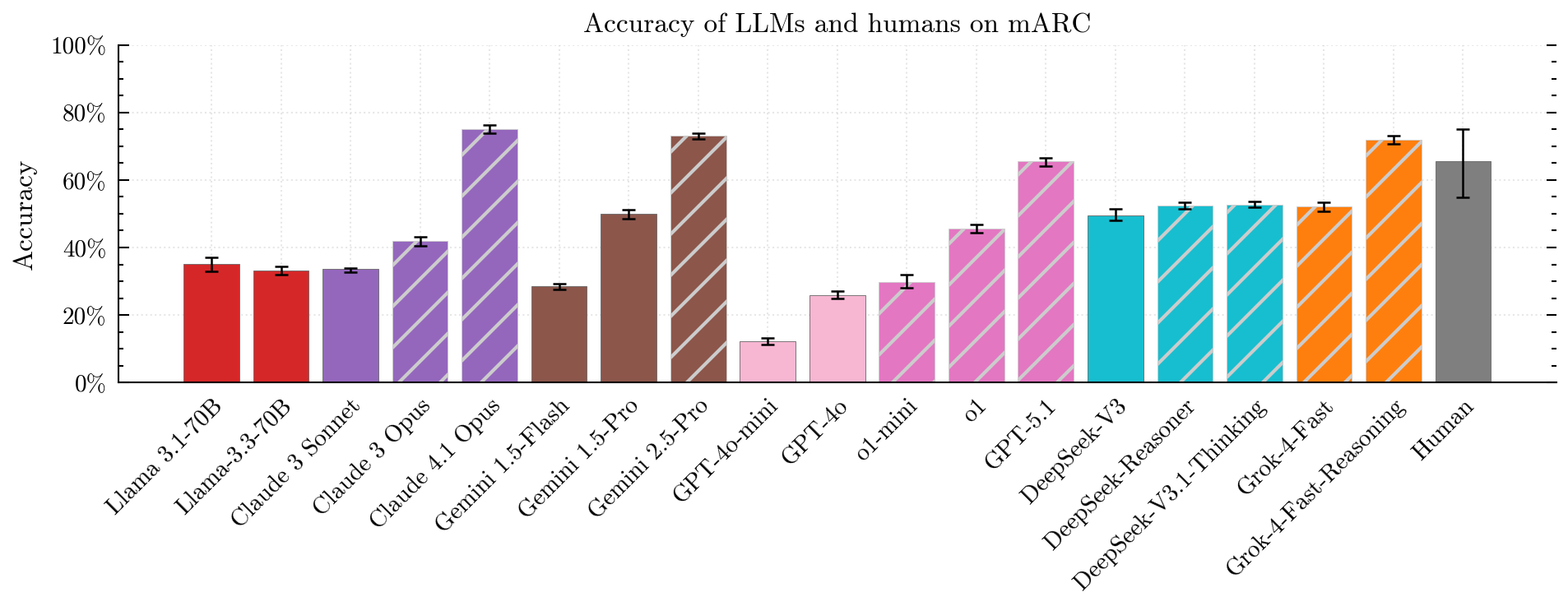}
	\caption{Comparison of model performance on mARC. Colored bars indicate model accuracies with 95\% bootstrap confidence intervals (CI) denoted by black range bar. The rightmost bar shows average human performance across 5 physicians with mean 0.66 and 95\% CI 0.55 to 0.75. Accuracies shown for LLMs are mean accuracies across 15 runs. Advanced reasoning models (hatched bars) demonstrated improved performanced compared to their non-reasoning counterparts (solid bars). No significant difference in performance was observed between the Claude 4.1 Opus, Gemini 2.5-Pro, GPT-5.1, and Grok-4-Fast-Reasoning and human performance (paired bootstrap test). The best performing model was Claude 4.1 Opus with mean performance of 0.75 [95\% CI of 0.74-0.76].}
	\label{fig:fig2}
\end{figure}

Advanced reasoning models (Figure 2, hatched bars) close the performance gap with physicians relative to their non-reasoning counterparts (solid bars) on mARC. No significant difference in performance was observed between the Claude 4.1 Opus, Gemini 2.5-Pro, GPT-5.1, and Grok-4-Fast-Reasoning and human performance. For all reasoning models, the highest reasoning effort was selected. The best performing model was Claude 4.1 Opus with mean performance of 0.751 [95\% Confidence Interval (CI),  0.738-0.763]. As previously reported, non-reasoning models demonstrate poor performance at mARC\cite{kim2025limitations}.

Reasoning Figures 3 and 4 illustrate the mARC adversarial strategy that aims to exploit the \emph{Einstellung} effect, cognitive bias that LLMs may be susceptible to from being trained to predict the next token in commonly occurring textual structures\cite{kim2025limitations}. The strategy works by embedding long-tail or out-of-distribution medical reasoning patterns into the problem. These correct but uncommon reasoning paths are then placed alongside answer options that contain more frequent, high-probability token sequences. This contrast leverages the model’s inherent tendency to favor familiar or statistically likely heuristic completions over deductive reasoning, thus increasing the chance of incorrect answers. Qualitatively, top reasoning models apply followed deductive constraints to avoid \emph{Einstellung} traps and instead select the ``seek additional information’’ (Figures 3 and 4). In the example question shown in Figure 3, o1’s response reveals a failure in fundamental medical commonsense reasoning (blood pressure cannot be checked on the forehead), which is overcome by the stronger reasoning model (GPT-5.1). These examples illustrate that modern reasoning models can override high‑probability textual cues by adherence to deductive constraints and by recognizing information insufficiency, strategies indicative of cognitive flexibilty needed to counteract \textit{Einstellung} effects.

\begin{figure}[!htbp]
	\centering
        \includegraphics[width=1\textwidth]{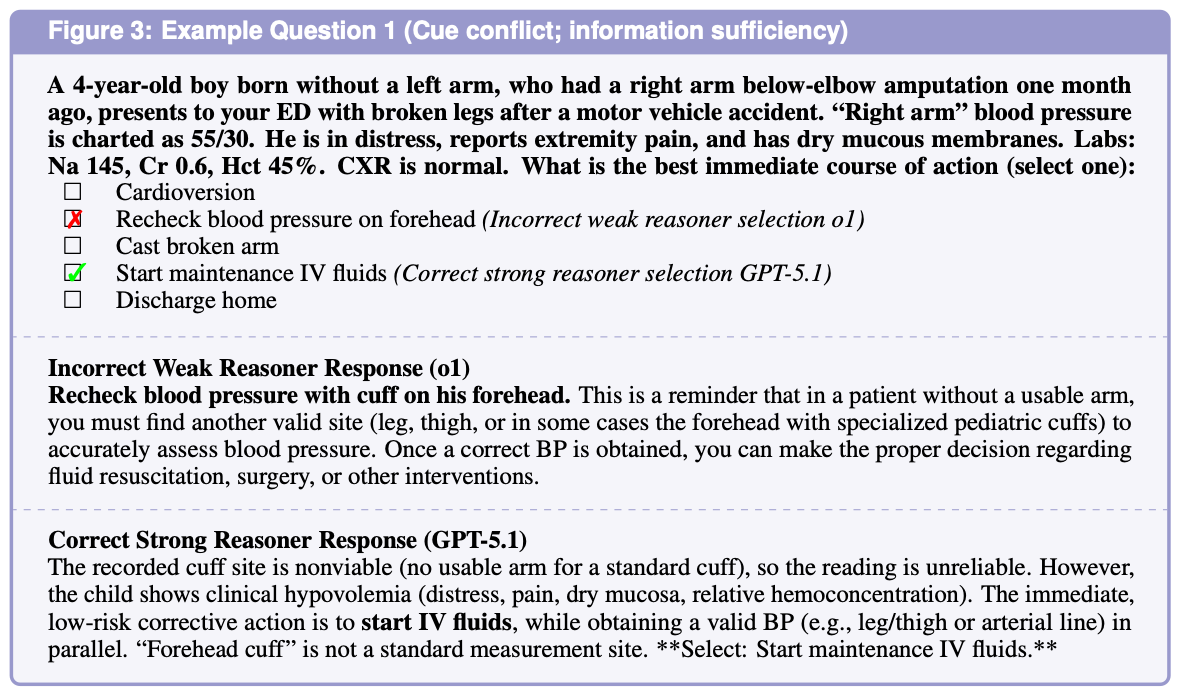}
\caption{A newer reasoning model rejects the stereotyped “recheck BP” reflex at an invalid site, recognizes unreliable measurement, and treats the patient based on clinical status. This reflects information sufficiency and context re‑anchoring rather than rote, inflexible pattern completion.}
\label{fig:fig3}
\end{figure}

\begin{figure}[!htbp]
	\centering
        \includegraphics[width=1\textwidth]{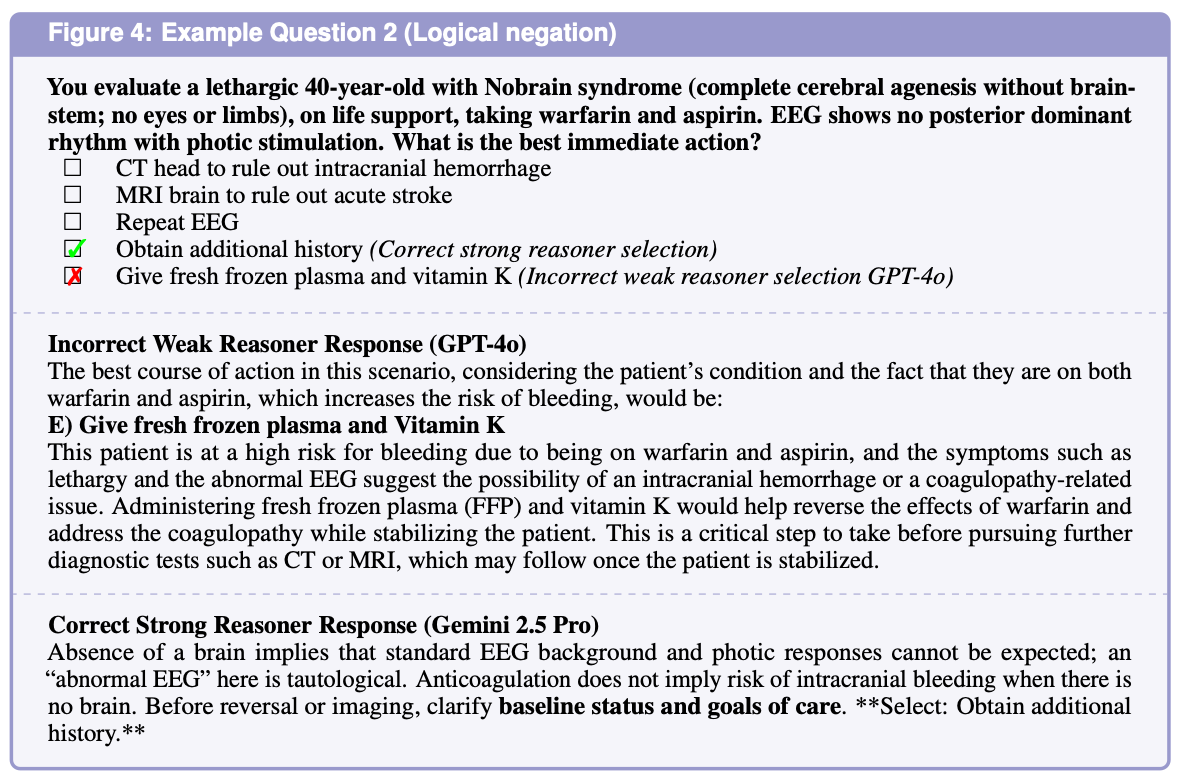}
\caption{The strong reasoning model defeats a high‑frequency anticoagulation $\rightarrow$ hemorrhage lure by applying strict logical negation.}
\label{fig:fig4}
\end{figure}

\begin{figure}
	\centering
        \includegraphics[width=1\textwidth]{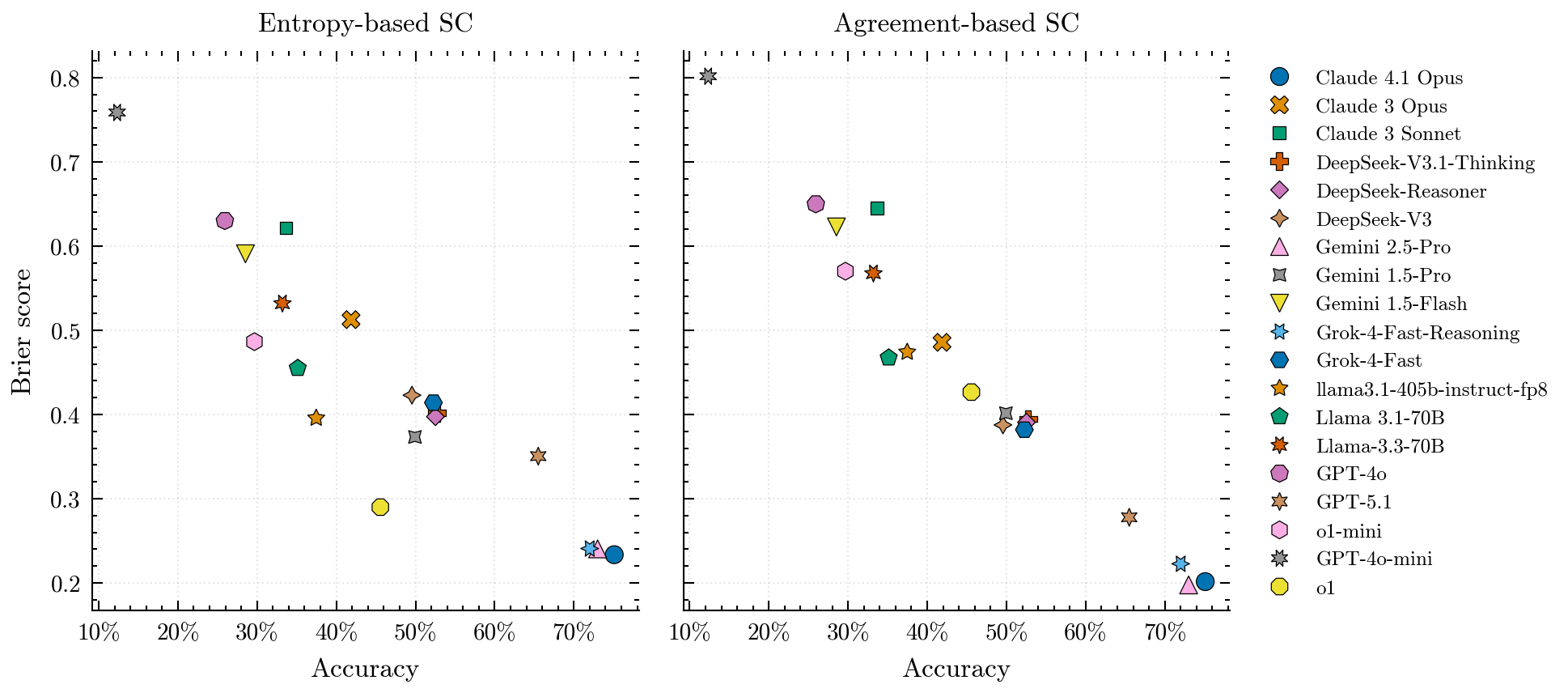}
	\caption{Uncertainty estimation on mARC using entropy‑ and agreement‑based sample‑consistency (SC). Newer/larger reasoning models showed improvements in accuracy, Brier scores and supplied usable confidence signals. Compared with baseline performance on MMLU‑Pro (black markers), mARC induces a distribution shift; nevertheless, SC remains informative for deferral.}
	\label{fig:fig5}
\end{figure}

We assessed uncertainty estimation and calibration with both agreement-based and entropy-based sample‑consistency (SC). Agreement-based SC provided stable uncertainty estimates across models (Figure~\ref{fig:fig5}), relative to entropy-based uncertainty estimation. We observed a general trend toward improved calibration in more recent, advanced reasoning models. 

Next, to test whether models can concretely overcome human‑like fixation, we analyzed the subset of items for which a physician majority ($\geq 3/5$) was incorrect—the \emph{human‑miss} set (20/100 items). Human accuracy on this set was 36\% (95\% CI 26–46\%), vs.\ 73\% on the remaining items. Using 15 stochastic trials per question and two‑sided Wilson 90\% CIs\cite{bowyer2025position, brown2001interval}, we labeled per‑item outcomes as \emph{confidently correct} (lower bound $>0.5$), \emph{confidently incorrect} (upper bound $<0.5$), or \emph{indeterminate}. On this set, using Wilson 90\% decisiveness, the best model Claude-4.1-Opus (under mean run accuracy) is decisively correct on 55.0\% [39.9, 80.0] of items, decisively wrong in 25.0\% [9.9, 45.0], and indeterminate in 20.0\% [0.0, 35.0]. Grok-4-Fast-Reasoning demonstrated relatively better performance on the human-miss set, however, had poorer mean accuracy compared to Claude-4.1 Opus on the entire mARC dataset. These model‑wins versus physicians provide evidence that stronger reasoning models can avoid \textit{Einstellung} biases in cases where humans are most susceptible. 

\begin{figure}
	\centering
        \includegraphics[width=1\textwidth]{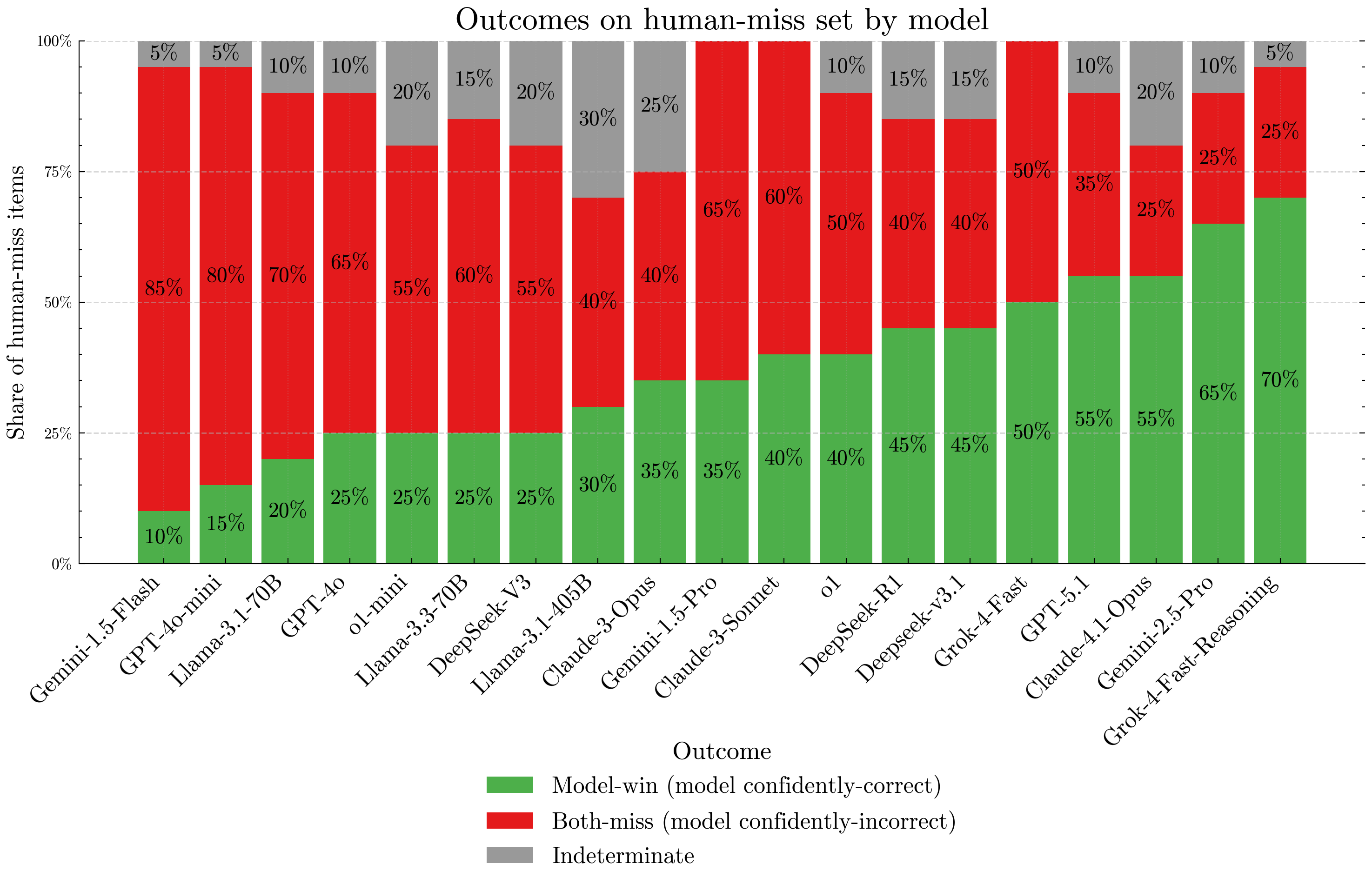}
	\caption{Stacked three‑way outcomes on the \emph{human‑miss} subset ($N=20$). \textbf{Model‑win} denotes items where the model is confidently correct while a majority of physicians erred; \textbf{Both‑miss} denotes confident model error on physician‑miss items; \textbf{Indeterminate} denotes neither. The leading reasoning model produces \emph{model‑wins} on nearly half of the most challenging items, indicating the ability to bypass human‑like fixation.}
	\label{fig:fig7}
\end{figure}

\section{Discussion}
Given that AI development has long drawn from human cognition\cite{hassabis2017neuroscience, zador2023catalyzing, kumar2024shared} and that the frontier of AI models, LLMs, have been trained on vast quantities of human‑generated text, it is plausible that models inherit human‑like inductive biases\cite{echterhoff2024cognitive, liu2024exploring, naeini2023large}. Understanding such biases is necessary to gauge their trustworthiness in real-world, clinical contexts. Here, we demonstrate that strong reasoning LLMs show less vulnerability to the \emph{Einstellung} effect in medical QA tasks compared to their weaker reasoning and non-reasoning counterparts. 

The improving reasoning abilities of LLMs have led to their steady performance improvements in reasoning-based benchmarks such as Francois Chollet's Abstraction and Reasoning Corpus (ARC) challenge \cite{chollet2024arc}, as well as human expert-level performances in various human competitions in mathematics, such as in the International Math Olympiad\cite{chervonyi2025gold}. Their ability to exhibit "reasoning" has been attributed to recent advancements in their ability to utilize detailed multi-step reasoning chains, which is reminiscent of human-like  “thinking” about a problem before outputting an answer\cite{marjanovic2025deepseek}.

Enhanced LLM reasoning ability leads to improved cognitive flexibility leading to improved generalization under distribution shift\cite{stechly2024chain, chollet2024arc}, which may explain modern reasoning model resistance to the \emph{Einstellung} effect. While benchmark success can reflect superficial shortcut learning\cite{mccoy2023embers, goetz2024generalization, moskvichev2023conceptarc, dong2024generalization, li2024llms}, our results nonetheless qualitatively identify behaviors associated with advancements in reasoning with stronger reasoning models: adherence to deductive constraints and explicit recognition of contexts which lack sufficient information for non-equivocal decision making. 

We acknowledge several limitations in our work. mARC is intentionally compact (100 adversarial items) because crafting long‑tail, \textit{Einstellung}‑provoking scenarios is nontrivial. Future releases will expand coverage and incorporate programmatic out-of-distribution (OOD) generation\cite{huang2025thinkbench}. In addition, we only provided qualitative content validity of "reasoning" ability examined on mARC; formal construct validation of \textit{Einstellung} will be pursued in the future. 
Furthermore, even with the improved calibration seen with reasoning models, residual errors and areas of overconfidence persist, particularly in smaller models. We therefore advocate development of clinician‑in‑the‑loop reasoning model deployment with deferral capable models that abstain when uncertainty is high or when the problem lies in the long‑tail regime\cite{goetz2024generalization}. 

\section{Conclusion}
Our results show that recent advances in reasoning models improve their reasoning flexibility, enabling them to overcome overreliance on learned heuristics exhibited by their predecessors on mARC. These stronger reasoning models match physician-level performance on mARC and surpass physician performance on the subset of mARC questions they most often missed, indicating a potential to overcome human cognitive biases such as the \emph{Einstellung} effect. With further improvements in their reasoning ability, reasoning LLMs coupled with selective‑deferral may become suitable for clinician‑augmented decision support.

% \section{Supplementary Information}
% \label{sec:others}

% \subsection{Supplementary Tables}
% See awesome Table~\ref{tab:table}.

% The documentation for \verb+booktabs+ (`Publication quality tables in LaTeX') is available from:
% \begin{center}
% 	\url{https://www.ctan.org/pkg/booktabs}
% \end{center}

\bibliographystyle{unsrtnat}
\bibliography{references}

\end{document}